\documentclass{article}

    \usepackage[preprint]{neurips_2020}

\usepackage{amsmath,amsfonts,bm}

\def\eqref#1{equation~\ref{#1}}

\def\floor#1{\lfloor #1 \rfloor}
\def\1{\bm{1}}

\DeclareMathAlphabet{\mathsfit}{\encodingdefault}{\sfdefault}{m}{sl}
\SetMathAlphabet{\mathsfit}{bold}{\encodingdefault}{\sfdefault}{bx}{n}

\usepackage{color}
\usepackage[usenames,dvipsnames]{xcolor}
\definecolor{shadecolor}{gray}{0.9}

\usepackage{hyperref}
\usepackage{url}
\usepackage[utf8]{inputenc} 
\usepackage[T1]{fontenc}    
\usepackage{hyperref}       
\usepackage{url}            
\usepackage{booktabs}       
\usepackage{amsfonts}       
\usepackage{nicefrac}       
\usepackage{microtype}      

\usepackage{times}
\usepackage[]{natbib}
\usepackage[toc,page]{appendix}
\usepackage{graphicx}
\usepackage{amsmath}
\usepackage{amssymb}
\usepackage{algorithm}
\usepackage[noend]{algpseudocode}
\usepackage{ragged2e}
\usepackage[nameinlink]{cleveref}
\creflabelformat{equation}{#2\textup{#1}#3}  
\usepackage{textcomp}
\usepackage{pgfplots}
\usepackage{colortbl}
\usepackage{wrapfig}
\usepackage{framed}
\usepackage{mathtools}

\usepackage{ragged2e}
\DeclareRobustCommand{\sidenote}[1]{\marginpar{
                                    \RaggedRight
                                    \textcolor{Plum}{\textsf{#1}}}}
\DeclareTextFontCommand{\emph}{\em}

\title{Measuring Calibration in Deep Learning}

\author{%
  Jeremy Nixon\thanks{Contact: jeremynixon@google.com.} \\
  Google Brain\\
   \And
   Mike Dusenberry \\
   Google Brain \\
   \And
   Ghassen Jerfel \\
   Google Brain \\
   \And
   Timothy Nguyen \\
   Google Research \\
   \And
   Jeremiah Liu \\
   Google Research \\
    \And
   Linchuan Zhang \\
   Google Inc. \\
   \And
   Dustin Tran \\
   Google Brain \\
}
\begin{document}

\maketitle

\begin{abstract}
Overconfidence and underconfidence in machine learning classifiers is measured by \emph{calibration}: the degree to which the probabilities predicted for each class match the accuracy of the classifier on that prediction.

How one measures calibration remains a challenge: expected calibration error, the most popular metric, has numerous flaws which we outline, and there is no clear empirical understanding of how its choices affect conclusions in practice, and what recommendations there are to counteract its flaws.

In this paper, we perform a comprehensive empirical study of choices in calibration measures including measuring all probabilities rather than just the maximum prediction, thresholding probability values, class conditionality, number of bins, bins that are adaptive to the datapoint density, and the norm used to compare accuracies to confidences. To analyze the sensitivity of calibration measures, we study the impact of optimizing directly for each variant with recalibration techniques. Across MNIST, Fashion MNIST, CIFAR-10/100, and ImageNet, we find that conclusions on the rank ordering of recalibration methods is drastically impacted by the choice of calibration measure. We find that conditioning on the class leads to more effective calibration evaluations, and that using the L2 norm rather than the L1 norm improves both optimization for calibration metrics and the rank correlation measuring metric consistency. Adaptive binning schemes lead to more stablity of metric rank ordering when the number of bins vary, and is also recommended. We open source a library for the use of our calibration measures.
\end{abstract}

\section{Introduction}

The reliability of a machine learning model's confidence in its predictions is critical for high risk applications, such as deciding whether to trust a medical diagnosis prediction \citep{crowson2016assessing,jiang2011calibrating,raghu2018direct,dusenberry2020}.
One mathematical formulation of the reliability of confidence is calibration \citep{murphy1967verification,dawid1982well}.

Intuitively, for class predictions, calibration means that if a model assigns a class with 90\% probability, that class should appear 90\% of the time.

Calibration is not directly measured by proper scoring rules like negative log likelihood or the quadratic loss. The reason that we need to assess the quality of calibrated probabilities is that presently, we optimize against a proper scoring rule and our models often yield uncalibrated probabilities.

Recent work proposed Expected Calibration Error \citep[ECE; ][]{naeini2015obtaining}, a measure of calibration error which has lead to a surge of works developing methods for calibrated deep neural networks \citep[e.g., ][]{guo2017calibration,kuleshov2018accurate}.
In this paper, we show that ECE has numerous pathologies, and that recent calibration methods, which have been shown to successfully recalibrate models according to ECE, cannot be properly evaluated via ECE.

Calibration (and uncertainty quantification generally) is critical in autonomous vehicles, the exploration phase of may algorithms, medical applications, and many more safety-critical applications of machine learning. A suite of recent papers \citep[e.g., ][]{lee2017training, vaicenavicius2019evaluating, thulasidasan2019mixup, kumar2018trainable, guo2017calibration, seo2019learning} use ECE to validate their models’ calibration. We identify concerns with that methodology. As this metric has become the default choice for measuring calibration in industry and for researchers, we expect our criticism to change the decisions of anyone attempting to train well-calibrated classification models.

\textbf{Contributions.} We study metrics for evaluating calibration. We make the following contributions: \begin{enumerate}
\item We identify problems with the widely used ECE metric to assess calibration. Specifically, problems with failing to condition on the class, the use of the L1 norm, and failing to assess all the predictions a model makes. 
\item We provide analysis of the major properties of calibration metrics, through experiments with rank correlation, class conditional recalibration, optimization for calibration error, metric consistency and rank correlation, and metric rank order sensivity to the binning hyperparameter.
\item We propose new metrics to address these problems.
\item We recommend best practices - conditioning on class, adaptivity and using the L2 norm - for evaluating calibration using our metrics.  
\end{enumerate}

\section{Background \& Related Work}

\subsection{Measurement of Calibration}
Assume the dataset of features and outcomes $\{(x, y)\}$ are i.i.d. realizations of the random variables $X,Y\sim \mathbb{P}$. We focus on class predictions. Suppose a model predicts a class $y$ with probability $\hat p$. The model is \emph{calibrated} if $\hat p$ is always the true probability.
Formally,
\begin{equation*}
\mathbb{P}(Y=y\mid \hat p=p) = p
\end{equation*}
for all probability values $p\in[0,1]$ and class labels $y\in\{0,\ldots,K-1\}$. The left-hand-side denotes the true data distribution's probability of a label given that the model predicts $\hat{p}=p$; the right-hand-side denotes that value.  Any difference between the left and right sides for a given $p$ is known as \textit{calibration error}.

\paragraph{Expected Calibration Error (ECE).}
To approximate the calibration error in expectation, ECE discretizes the probability interval into a fixed number of bins, and assigns each predicted probability to the bin that encompasses it.
The calibration error is the difference between the fraction of predictions in the bin that are correct (accuracy) and the mean of the probabilities in the bin (confidence). Intuitively, the accuracy estimates $\mathbb{P}(Y=y\mid \hat p=p)$, and the average confidence is a setting of $p$. ECE computes a weighted average of this error across bins:
\begin{equation*}
\operatorname{ECE} = \sum_{b=1}^{B} \frac{n_b}{N} \left| \operatorname{acc}(b) - \operatorname{conf}(b) \right|,
\end{equation*}
where $n_b$ is the number of predictions in bin $b$, $N$ is the total number of data points, and $\operatorname{acc}(b)$ and $\operatorname{conf}(b)$ are the accuracy and confidence of bin $b$, respectively.
ECE as framed in \citet{naeini2015obtaining} leaves ambiguity in both its binning implementation and how to compute calibration for multiple classes. In \citet{guo2017calibration}, they bin the probability interval $[0,1]$ into equally spaced subintervals,
and they take the maximum probability output for each datapoint (i.e., the predicted class's probability). We use this for our ECE implementation.

Adaptivity was explored by \citep{nguyen2015posterior} and applied in \citep{hendrycks2018deep}.
\citet{kuleshov2018accurate} extends ECE to the regression setting.
Unlike the Brier score \citep{brier1950} and Hosmer-Lemeshow \citep{hosmer1980goodness}, we'd like the metric to be scalar-valued in order to easily benchmark methods, and to only measure calibration.




\subsection{Proper Scoring Rules \& Other Measures of Calibration}
Many classic methods exist to measure the accuracy of predicted probabilities.
For example, the Brier score (quadratic loss) measures the mean squared difference between the predicted probability and the actual outcome \citep{brier1950, gneiting2007strictly}.
This score can be shown to decompose into a sum of metrics, including calibration error. In practice, training against this proper scoring rule, like training against log likelihood, does not guarantee a well calibrated model.
The Hosmer-Lemeshow test is a popular hypothesis test for assessing whether a model's predictions significantly deviate from perfect calibration \citep{hosmer1980goodness}.
%
%
The reliability diagram provides a visualization of how well-calibrated a model is \citep{degroot1983comparison}.

\subsection{Recalibration Methods}
\label{sec:recalibration}

A common approach to calibration is to apply post-processing methods to the output of a classifier without retraining. The two most popular post-processing methods are the parametric approach of Platt scaling \citep{platt1999probabilistic} and the non-parametric approach of isotonic regression \citep{zadrozny2002transforming}.

\textbf{Platt scaling} fits a logistic regression model to the logits of a classifier, on the validation set, which can be used to compute calibrated predictions at test time. The original formulation of Platt scaling for neural networks \citep{niculescu2005predicting} involves learning scalar parameters $a,b \in \mathbb{R}$ on a held-out validation set and then computing calibrated probabilities $\hat{p_{i}}$ given the uncalibrated logits vector $z_i$ on the test set as $\hat{p_{i}} = \sigma(az_i + b)$. These parameters are typically estimated by minimizing the negative log likelihood.

Platt scaling can be extended to the multiclass setting by considering higher-dimensional parameters. In \textbf{matrix scaling} $a$ is replaced with $W \in \mathbb{R}^{K*K}$ while we consider $b \in \mathbb{R}^{K}$. As for \textbf{vector scaling}, $W \in \mathbb{R}^{K}$. Calibrated probabilities for either of these extensions can then be computed as
$$\hat{p_{i}} = \max_{k} \sigma(Wz_i + b).$$
An even simpler extension is \textbf{temperature scaling} \citep{guo2017calibration} which reduces the set of regression parameters to the inverse of a single scalar $T > 0$ such that
$$\hat{p_{i}} = \max_{k} \sigma(z_i/T).$$
\textbf{Isotonic regression} is a common non-parametric processing method that finds the stepwise-constant non-decreasing (isotonic) function $f$ that best fits the data according to a mean-squared loss function $\sum_i (f(p_i) - y_i)^2$ where $p_i$ are the uncalibrated probabilities and $y_i$ the labels. 

The standard approach for extending isotonic regression to the multiclass setting is to break the problem into many binary classification problems (e.g. one-versus-all problems), to calibrate each problem separately, and then to combine the calibrated probabilities \citep{zadrozny2002transforming}.

\section {Issues With Calibration Metrics}
\subsection{Not Computing Calibration Across All Predictions}

%
Expected Calibration Error was crafted to mirror reliability diagrams, which are structured around binary classification such as rain vs not rain \citep{degroot1983comparison} (Figure 1). A consequence is that the error metric is reductive in a multi-class setting.
In particular, ECE is computed using only the predicted class's probability, which implies the metric does not assess how accurate a model is with respect to the $K-1$ other probabilities.

We examine increasing the importance of omitted predictions via label noise. As label noise increases, trained models become less confident in their predictions. Secondary and tertiary predictions are correct with higher frequency, and so for many samples the prediction with the correct label isn't included in ECE's measure of calibration error. We find that ECE becomes a worse approximation of the calibration error as relevant predictions not included in the ECE metric become common (Figure 1).

\begin{figure}
\includegraphics[width=.49\columnwidth]{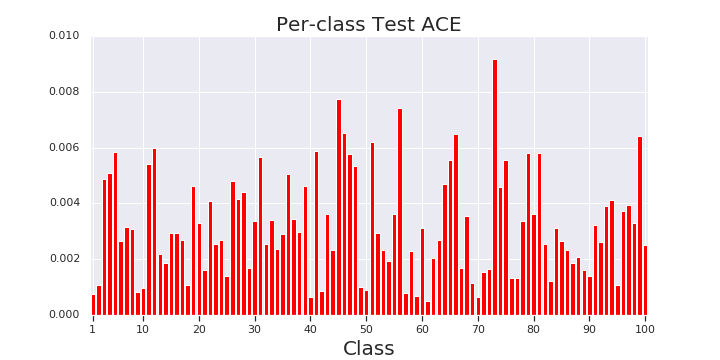}
\includegraphics[width=.35\columnwidth]{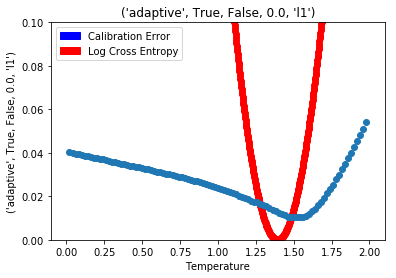}
\caption{\textbf{Left:} Per-class Adaptive Calibration Error (ACE) for a trained 110-layer ResNet on the CIFAR-100 test set. We observe that calibration error is non-uniform across classes, which is difficult to express
when measuring error only on the maximum probabilities. \textbf{Right:} Calibration loss surface as the temperature changes. Notice that the temperature that minimizes the calibration error is not the same as the temperature that minimizes the cross entropy loss function.}

\label{fig:resnet-cifar100-per-class-metrics-tace}

\end{figure}

In \Cref{fig:resnet-cifar100-per-class-metrics-tace}, we observe that calibration error, when measured in a per-class manner, is non-uniform across classes. Guaranteeing calibration for all class probabilities is an important property for many applications. For example, a physician may care deeply about a secondary or tertiary prediction of a disease. Decision costs along both ethical and monetary axes are non-uniform, and thus clinical decision policies necessitate a comprehensive view of the likelihoods of potential outcomes \citep[e.g., ][]{tsoukalas2015}.  Alternatively, if a true data distribution exhibits high data noise (also known as high aleatoric uncertainty, or weak labels), then we'd like models to be calibrated across all predictions as no true class label is likely.

\subsection{Fixed Calibration Ranges}
One major weakness of evenly spaced binning metrics is caused by the dispersion of data across ranges. In computing ECE, there is often a large leftward skew in the output probabilities, with the left end of the region being sparsely populated and the rightward end being densely populated. Sharpness, which is the desire for models to always predict with high confidence, i.e., predicted probabilities concentrate to 0 or 1, is a fundamental property \citep{gneiting2007probabilistic}. (That is, network predictions are typically very confident.) This causes only a few bins to contribute the most to ECE---typically one or two as bin sizes are 10-20 in practice \citep{guo2017calibration}.



\subsection{Bias-Variance Tradeoff}



Selecting the number of bins has a bias-variance tradeoff as it determines how many data points fall into each bin and therefore the quality of the estimate of calibration from that bin's range. In particular, a larger number of bins causes more granular measures of calibration error (low bias) but also a high variance of each bin's measurement as bins become sparsely populated. This tradeoff compounds particularly with the problem of fixed calibration ranges, as due to sharpness certain bins have many more data points than others.

\subsection{Pathologies in Static Binning Schemes}
Metrics that depend on static binning schemes like ECE
suffer from issues where you can get near 0 calibration error due to overconfident and underconfident predictions overlapping in the same bin. These cancellation effects may not always be extreme enough to push an incredibly uncalibrated model to 0 calibration error, but it can be difficult to judge when using a static metric whether differences between models or techniques are due to a true improvement in the model's calibration or whether improved calibration is due to cancellation. 



\section{Analyzing The Space of Calibration with General Calibration Error}

To study choices in measuring calibration, we examine General Calibration Error (GCE), an error metric captures the space of calibration metrics. It takes various inputs for the following five properties as arguments and returns the corresponding metric.

\subsection{Class Conditionality}

Rather than lumping probabilities for each class together into a single vector to sort and bin, and only then computing the calibration error, the calibration error can be computed over the class vectors independently of one another, and only be averaged afterwards. This allows systematic differences in the calibration error between classes to be evaluated without washing one another out. Class weighting is even when all datapoints are considered in the loss, but when focusing on the maximum probabilities or under thresholding the class weighting skews towards the classes on which the model places the most confidence.

\subsection{Adaptivity}

Adaptive calibration ranges are motivated by the bias-variance tradeoff in the choice of ranges, suggesting that in order to get the best estimate of the overall calibration error the metric should focus on the regions where the predictions are made (and focus less on regions with few predictions).  This leads us to introduce Adaptive Calibration Error (ACE).
uses an adaptive scheme which spaces the bin intervals so that each contains an equal number of predictions.

In detail, ACE takes as input the predictions $P$ (usually out of a softmax), correct labels, and a number of ranges $R$.

\begin{equation*}
\operatorname{ACE} = \frac{1}{KR}\sum_{k=1}^K \sum_{r=1}^{R} \left| \operatorname{acc}(r, k) - \operatorname{conf}(r, k) \right|.
\end{equation*}
Here, $\operatorname{acc}(r, k)$ and $\operatorname{conf}(r, k)$ are the accuracy and confidence of adaptive calibration range $r$ for class label $k$, respectively; and $N$ is the total number of data points. Calibration range $r$ defined by the $\floor{N/R}$th index of the sorted and thresholded predictions.

\subsection{Maximum Probability}



The most popular calibration measure, ECE, does not measure the calibration of all probabilities - it focuses only on the probability that have the maximum probability for a given datapoint. This approach made sense in the binary setting. We first introduce Static Calibration Error (SCE), which is a simple extension of Expected Calibration Error to every probability in the multiclass setting. SCE bins predictions separately for each class probability, computes the calibration error within the bin, and averages across bins:
\begin{equation*}
\operatorname{SCE} = \frac{1}{K} \sum_{k=1}^K \sum_{b=1}^{B} \frac{n_{bk}}{N} \left| \operatorname{acc}(b, k) - \operatorname{conf}(b, k) \right|.
\end{equation*}
Here, $\operatorname{acc}(b, k)$ and $\operatorname{conf}(b, k)$ are the accuracy and confidence of bin $b$ for class label $k$, respectively; $n_{bk}$ is the number of predictions in bin $b$ for class label $k$; and $N$ is the total number of data points. Unlike ECE, assuming infinite data and infinite bins, SCE is guaranteed to be zero if only if the model is calibrated.


\subsection{Norm}

The norm chosen to compare accuracy against confidence in the General Calibration Error can be either the L1 norm $\sum|acc(b, k) - conf(b,k)|$ or the L2 norm $\sqrt{\sum(acc(b, k) - conf(b, k))^2}$. 

\subsection{Thresholding}

One challenge is that the vast majority of softmax
predictions become infinitesimal. These tiny predictions
can wash out the calibration score, especially in the case
where there are many classes, where a large proportion
of them model’s predictions correspond to an incorrect
class. This leads to a very inefficient use of bins, where a number of bins close to the number of classes minus one will
be devoted to unlikely regions. One natural solution to that challenge is thresholding -
only including predictions above some epsilon (ex., .01
or .001) in the calibration score. The difference between this approach and focusing on the maximum probability for each datapoint is the large number of secondary and tertiary predictions of substiantial value (up to 49.9\%) that are rejected when a measure focuses on the maximum probability.

\section{Experiments}

\subsection{Class Conditionality}

Measuring histogram binning accuracy via class conditional metrics requires a reevaluation of its effectiveness. Because histogram binning optimizes directly for a notion of the calibration error, it will be overconfidently measured by a measure of calibration error that exactly matches its assumptions. Specifically, we compare class conditional histogram binning to class unconditional histogram binning. In class conditional histogram binning, probabilities come from the average accuracy on the validation dataset from the given classes' binned probability, rather than conflating and bundling probabilities from many classes together before binning as in standard histogram binning.

We find that on CIFAR 10 the ECE estimate of the calibration error is one-third that of the class conditional variant, which is the same as ECE in all respects but class conditionality (\Cref{tab:class_conditionality_cifar10}). On Imagenet  (\Cref{tab:class_conditionality_imagenet}) we find that the ECE for histogram binning is one-fifth of the class-conditional measure after controlling using the uncalibrated score.

\begin{table}
\begin{center}
\begin{tabular}{l|llll}
\textbf{CIFAR 10}  &  \textbf{Uncalibrated} &  \textbf{Histogram Binning}      &  \textbf{Class Conditional Histogram Binning}          \\
\hline
\begin{tabular}[c]{@{}l@{}}ECE\end{tabular}        &  0.02447  &    0.0070 &    0.0132            \\ 
\begin{tabular}[c]{@{}l@{}}Class Conditional ECE\end{tabular}       &   0.03011 &    0.0212    &    0.0189       \\
\end{tabular}
\end{center}
\caption{ECE overestimates Histogram Binning's calibraiton error.}
\label{tab:class_conditionality_cifar10}
\end{table}

\begin{table}
\begin{center}
\begin{tabular}{l|llll}
\textbf{Imagenet}  &  \textbf{Uncalibrated} &  \textbf{Histogram Binning}      &  \textbf{Class Conditional Histogram Binning}          \\
\hline
\begin{tabular}[c]{@{}l@{}}ECE\end{tabular}    &  0.0669  & 0.0147 & 0.0831    \\ 
\begin{tabular}[c]{@{}l@{}}Class Conditional ECE\end{tabular}    &  0.2237  &  0.1810   &  0.1561 \\
\end{tabular}
\end{center}
\caption{ECE overestimates Histogram Binning's calibraiton error.}
\label{tab:class_conditionality_imagenet}
\end{table}

\subsection{Optimizing for Calibration Error}

We get evidence of the impact of each property by optimizing the temperature scaling parameter for the calibration error of our metrics. An example of one of these surfaces is \Cref{fig:resnet-cifar100-per-class-metrics-tace}. The largest differences are between the even and adaptive binning schemes, as well as the L1 and L2 norm - when the metric being optimized for is even, downstream performance on even metrics is strong (\Cref{tab:optimize_calibration_error}). ECE is very strongly influenced by measures of calibration error that adhere to its own properties, rather than capturing a more general concept of the calibration error. For example, ECE appears to show that on Imagenet, optimizing for the L1 Norm rather than the L2 norm leads to a halving of the calibration error, and shows an even more stark difference for being class unconditional.

Our model trained on Imagenet and CIFAR 10 in this experiment is a Wide Resnet. The optimizer used in these experiments is the Nelder-Mead downhill simplex algorithm.

\begin{table}
\begin{center}
\hspace*{-0.5cm}
\begin{tabular}{l|llll}
\textbf{Property} &  \textbf{Imagenet}      &  \textbf{CIFAR-10} &  \textbf{CIFAR-10 Control CE}\\
\hline
\begin{tabular}[c]{@{}l@{}}Even Binning Scheme\end{tabular}        &  0.0322        &    0.0097 & 0.0035 \\ 
\begin{tabular}[c]{@{}l@{}}Adaptive Binning Scheme\end{tabular}       &   0.0480 &   0.0175 & 0.0034\\
\hline
\begin{tabular}[c]{@{}l@{}}Max Probs True\end{tabular}      &  0.0377         &   0.0129 & 0.0035 \\
\begin{tabular}[c]{@{}l@{}}Max Probs False\end{tabular}      &   0.0426       &  0.0143 & 0.0034\\
\hline
\begin{tabular}[c]{@{}l@{}}Class Conditional\end{tabular}      &   0.0589        &  0.0145 & 0.0037 \\
\begin{tabular}[c]{@{}l@{}}Class Unconditional\end{tabular}      &   0.0213        &  0.0126 & 0.0033 \\
\hline
\begin{tabular}[c]{@{}l@{}}Thresholded\end{tabular}      &   0.0412        &  0.0126 & 0.0035\\
\begin{tabular}[c]{@{}l@{}}Unthresholded\end{tabular}        &  0.0390         &  0.0146 & 0.0034\\ 
\hline
\begin{tabular}[c]{@{}l@{}}L1 Norm\end{tabular}       &   0.0535 & 0.0117 & 0.0036 \\
\begin{tabular}[c]{@{}l@{}}L2 Norm\end{tabular}      &   0.0267  &  0.0155 & 0.0033\\
\hline
\end{tabular}
\end{center}
\caption{When optimizing for mutiple notions of the calibration error and evaluating using Expected Calibration Error, ECE is dominated by the similarity of the calibration measure to itself, rather than capturing a more general notion of calibration error. Here we report the mean ECE when optimizing for all metrics with a given property. The control group calibration error is RMSCE \citep{hendrycks2018deep}}
\label{tab:optimize_calibration_error}
\end{table}

\subsection{Sensitivity to the Number of Bins Hyperparameter}

One important hyperparemeter in each measure of calibration error is the number of bins to partition the data into. We analyse the mean rank-order differences between the combinatoric combinations of the same calibration measure evaluated with binning hyperparameters at values of 10, 20, 30, 40 and 50. We find dramatic differences in bin sensitivity depending on properties of the metric at hand. Adaptive calibration measures dramatically outperform evenly binned calibration measures in terms of mean rank correlation over the recalibration techniques over a Wide Residual Network trained on CIFAR 10. The rank correlation is 0 when two rankings are indpendent of one another, 1 when rankings are perfectly correlated, and -1 when rankings are inversely correlated.

Rank Correlation:
$$\rho = 1 - \frac{6 \sum_n |r_1 - r_2|}{n(n^2 - 1)}$$

Where $r_1$ and $r_2$ are the rankings being compared and $n$ is the number of values in the ranking.


The mean rank correlation for the adaptive binning metrics is $0.5927$, substiantially higher than the even binning measure of $0.3677$. This result leads us to recommend adapative binning as a method to improve rank order robustness to the number of bins hyperparameter. Other bin sensitivity results include $0.3565$ with max probs and $0.6037$ without, $0.5546$ with class conditionality and $0.4056$ without, $0.5010$ with thresholding and $0.4592$ without, and $0.5352$ with the L1 score and $0.4250$ without.

\begin{table}
\begin{center}
\hspace*{-0.5cm}
\begin{tabular}{l|llll}
\textbf{Property} &  \textbf{CIFAR-10}      &  \textbf{Fashion-MNIST}        &  \textbf{MNIST}   \\
\hline
\begin{tabular}[c]{@{}l@{}}Even Binning Scheme\end{tabular}        &  0.5775         &    0.2944            &  0.1569 \\ 
\begin{tabular}[c]{@{}l@{}}Adaptive Binning Scheme\end{tabular}       &   0.5084 &   0.2063           & 0.1156 \\
\hline
\begin{tabular}[c]{@{}l@{}}Max Probs True\end{tabular}      &  0.5018         &    0.2901   &  0.1652 \\
\begin{tabular}[c]{@{}l@{}}Max Probs False\end{tabular}      &   0.5911        &   0.2018           &  0.1059 \\
\hline
\begin{tabular}[c]{@{}l@{}}Class Conditional\end{tabular}      &   0.5162        &    0.1717           &  0.1941 \\
\begin{tabular}[c]{@{}l@{}}Class Unconditional\end{tabular}      &   0.6156        &   0.2275           &  0.1229 \\
\hline
\begin{tabular}[c]{@{}l@{}}Thresholded\end{tabular}      &   0.4970        &   0.1994           &  0.1310 \\
\begin{tabular}[c]{@{}l@{}}Unthresholded\end{tabular}        &  0.5759         &   0.1710          & 0.1093 \\ 
\hline
\begin{tabular}[c]{@{}l@{}}L1 Norm\end{tabular}       &   0.4307&   0.2025            & 0.3475\\
\begin{tabular}[c]{@{}l@{}}L2 Norm\end{tabular}      &   0.6046         &    0.2720  &  0.1944 \\
\hline
\end{tabular}
\end{center}
\caption{We measure the mean rank correlation between orderings of the calibration metrics over our space of properties on 10 models per dataset, all trained with the same architecture. We expect that metrics would agree about which models were most calebrated, least calibrated, and so have a very similar ordering.}
\vspace{-2em}
\label{tab:rank_correlation}
\end{table}



\subsection{Recalibration Experiments}

\begin{table}
\begin{center}
\begin{tabular}{l|lllllllllllllllll}
\textbf{Rank Order} & \textbf{0 } & \textbf{1 } & \textbf{2 } & \textbf{3 } & \textbf{4 } & \textbf{5 } & \textbf{6 } & \textbf{7 } & \textbf{8 } & \textbf{9 } &  \textbf{10} & \textbf{11} & \textbf{12} & \textbf{13} & \textbf{14} & \textbf{15} & 
\\
\hline
\begin{tabular}[c]{@{}l@{}}Ranked First\end{tabular}  &  0 & 7 & 0 & 7 & 3 & 3 & 3 & 3 & 0 & 4 & 0 & 7 & 4 & 4 & 2 & 4 & \\ 
\begin{tabular}[c]{@{}l@{}}Ranked Second\end{tabular}  & 1 & 2 & 1 & 2 & 1 & 4 & 1 & 4 & 1 & 3 & 7 & 2 & 3 & 3 & 6 & 3

\\ 
\begin{tabular}[c]{@{}l@{}}Ranked Third\end{tabular}  & 3 & 3 & 3 & 3 & 4 & 1 & 4 & 1 & 4 & 7 & 1 & 6 & 1 & 1 & 3 & 7

\\ 
\begin{tabular}[c]{@{}l@{}}Ranked Fourth\end{tabular}  & 4 & 4 & 4 & 4 & 6 & 7 & 6 & 7 & 3 & 6 & 2 & 4 & 0 & 7 & 1 & 2

\\ 
\begin{tabular}[c]{@{}l@{}}Ranked Fifth\end{tabular}  & 2 & 6 & 2 & 6 & 2 & 2 & 2 & 2 & 7 & 0 & 6 & 3 & 7 & 0 & 7 & 1

\\ 
\begin{tabular}[c]{@{}l@{}}Ranked Sixth\end{tabular}  & 7 & 5 & 7 & 5 & 0 & 6 & 0 & 6 & 6 & 1 & 4 & 5 & 6 & 6 & 4 & 6

\\ 
\begin{tabular}[c]{@{}l@{}}Ranked Seventh\end{tabular}  & 5 & 1 & 5 & 1 & 7 & 0 & 7 & 0 & 5 & 5 & 3 & 1 & 5 & 5 & 0 & 0

\\ 
\begin{tabular}[c]{@{}l@{}}Ranked Eighth\end{tabular}  & 6 & 0 & 6 & 0 & 5 & 5 & 5 & 5 & 2 & 2 & 5 & 0 & 2 & 2 & 5 & 5
\\ 
\hline
\end{tabular}

\begin{tabular}{l|lllllllllllllllll}
\textbf{Rank Order} & \textbf{16} & \textbf{17} & \textbf{18} & \textbf{19} & \textbf{20} & \textbf{21} & \textbf{22} & \textbf{23} & \textbf{24} & \textbf{25} & \textbf{26} & \textbf{27} & \textbf{28} & \textbf{29} & \textbf{30} & \textbf{31}
\\
\hline
\begin{tabular}[c]{@{}l@{}}Ranked First\end{tabular}   & 1 & 2 & 1 & 2 & 1 & 1 & 1 & 1 & 0 & 0 & 0 & 7 & 1 & 1 & 1 & 1
\\
\begin{tabular}[c]{@{}l@{}}Ranked Second\end{tabular}   & 0 & 7 & 0 & 7 & 2 & 6 & 2 & 6 & 1 & 6 & 1 & 6 & 7 & 7 & 2 & 2

\\ 
\begin{tabular}[c]{@{}l@{}}Ranked Third\end{tabular}  & 2 & 6 & 2 & 6 & 7 & 2 & 7 & 2 & 3 & 7 & 2 & 2 & 0 & 6 & 7 & 7
\\ 
\begin{tabular}[c]{@{}l@{}}Ranked Fourth\end{tabular}  & 5 & 3 & 5 & 3 & 6 & 7 & 6 & 7 & 4 & 1 & 7 & 4 & 6 & 0 & 6 & 6

\\ 
\begin{tabular}[c]{@{}l@{}}Ranked Fifth\end{tabular}  & 7 & 4 & 7 & 4 & 3 & 3 & 3 & 3 & 7 & 3 & 6 & 3 & 3 & 2 & 0 & 4

\\ 
\begin{tabular}[c]{@{}l@{}}Ranked Sixth\end{tabular}  & 4 & 5 & 4 & 5 & 0 & 4 & 4 & 4 & 6 & 4 & 4 & 0 & 4 & 3 & 4 & 3

\\ 
\begin{tabular}[c]{@{}l@{}}Ranked Seventh\end{tabular}  & 3 & 0 & 3 & 0 & 4 & 5 & 0 & 5 & 5 & 5 & 3 & 5 & 5 & 4 & 3 & 0

\\ 
\begin{tabular}[c]{@{}l@{}}Ranked Eighth\end{tabular}  & 6 & 1 & 6 & 1 & 5 & 0 & 5 & 0 & 2 & 2 & 5 & 1 & 2 & 5 & 5 & 5
\\ 
\hline
\end{tabular}

\end{center}
\caption{Do different calibration error measures give the same ordering to a set of probabilities? Here, we measure the rank ordering of 8 recalibration techniques on CIFAR 10, expecting different measures of the calibration error to report the same ordering. Instead, we see that different calibration metrics report dramatically different orderings on the same set of predictions. Method numbers are as follows: 0: Class-Conditional Histogram Binning. 1: Boostrap Histogram Binning. 2: Isotonic Regression. 3: Temperature Sclaing. 4: ECE Optimized Temperature Scaling. 5: Vector Scaling. 6: Matrix Scaling. 7: Neural Network Scaling. Calibration metrics 0-31 are described in the supplemental material / appendix.}
\vspace{-1em}
\label{tab:recalibration_rank}
\end{table}

\textbf{Rank Ordering Across 32 Metrics.} More broadly outside of specific analysis for three metrics, we also examine the overall ranking of recalibration techniques across all possible metrics.

We observe dramatic inconsistency between the calibration ordering on models trained on the same dataset with the same architecture.
This demonstrates that varying the properties of the calibration error metric can lead to different conclusions about which model has the best calibration.

Do different calibration error measures give the same ordering to a set of probabilities? In \Cref{tab:recalibration_rank} we measure the rank ordering of each of our 32 metrics. ECE corresponds to $4$, Class Conditional ECE to $0$, SCE to $8$, ACE to $24$, RMSCE to $21$, and others whose index to metric mapping is documented in the appendix / supplement. The base data is recalibrated probabilities on CIFAR 10 over a wide residual network.


For example, temperature scaling \citep{guo2017calibration} has been shown to effectively minimize expected calibration error better than alternative techniques such as isotonic regression and Platt scaling \citep{platt1999probabilistic}. The question of whether these scaling techniques effectively minimize more sophisticated error metrics is an important standard for their efficacy.


When designing post-processing methods, not accounting for the properties detailed in Section 4 during evaluation can lead to misleading conclusions about a post-processing method's success.

\section{Discussion}
This paper studies metrics for evaluating calibration. We make the following contributions: \begin{enumerate}
\item We identify problems with the widely used ECE metric to assess calibration.  
\item We propose new metrics (SCE, ACE and GCE) to explore and ameliorate each of these problems.  
\item We recommend best practices for how to evaluate calibration.  
\end{enumerate}
We believe that the way neural network uncertainties are evaluated in future must be aware of the challenges we raise, and in light of those challenges recommend the use of GCE over ECE for evaluating a multiclass classifier's calibration error.

\textbf{Criticisms \& Limitations.}
Adaptive calibration metrics can create excessively large ranges if there is very sparse output in a region. In those cases, it may be better to have a higher variance estimate of the calibration, but only compare datapoints that are closer together (dropping datapoints that are farther away). This level of granularity can be replicated with an adaptive scheme that has a very large number of ranges, but that will come at the cost of easy interpretability and will increase the within-range variance across all ranges, not just the sparse regions.



\bibliography{neurips_2020.bib}
\bibliographystyle{neurips_2020.bst}

\clearpage

\appendix
\section{Appendix}
\subsection{Thresholding \& Thresholded Adaptive Calibration Error}
One initial challenge is that the vast majority of softmax predictions become infinitesimal (Figure 3, Top Left).
These tiny predictions can wash out the calibration score, especially in the case where there are many classes, where a large proportion of them model's predictions correspond to an incorrect class. 
One response is to only evaluate on values above a threshold $\epsilon$. Mathematically, TACE is identical to ACE, with the only difference being that TACE is only evaluated on values above $\epsilon$.
These predictions overlap with the predictions evaluated by ECE (all maximum values per datapoint), leading them to have similar reactions to recalibration methods.

\begin{figure}[!htb]
\includegraphics[width=0.49\columnwidth]{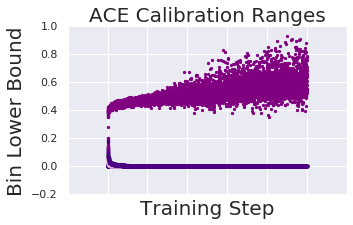}
\includegraphics[width=0.49\columnwidth]{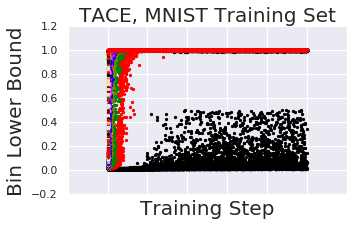}
\includegraphics[width=0.49\columnwidth]{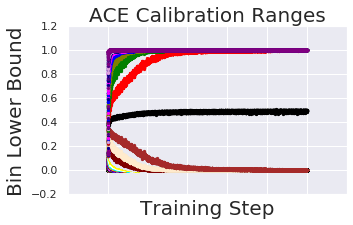}
\includegraphics[width=0.49\columnwidth]{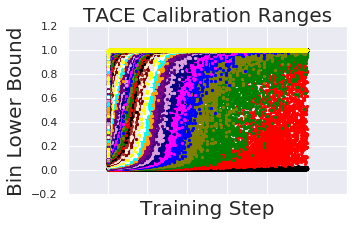}
\caption{\textbf{Top Left:} Lower bounds of calibrations ranges over the course of training for adaptive calibration error on Fashion-MNIST, focusing almost entirely on small ranges and motivating thresholding. \textbf{Top Right:} On the MNIST training set with thresholding, so few values are small that the bottom of the lowest range often spikes to .99 and higher due to every datapoint being fit. \textbf{Bottom Left:} ACE on Fashion-MNIST validation with 100 calibration ranges. \textbf{Bottom Right:} Thresholded adaptive calibration with 50 calibration ranges over the course of training on Fashion-MNIST's validation set.}
\label{fig:mnist-bin}
\end{figure}

\begin{figure}
\includegraphics[width=.49\columnwidth]{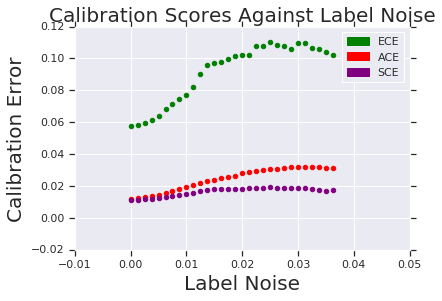}
\includegraphics[width=.49\columnwidth]{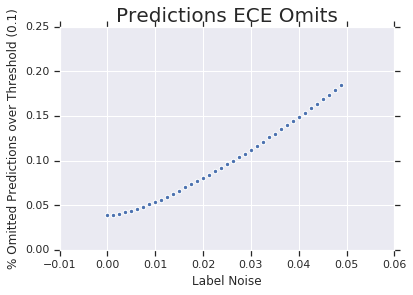}
\includegraphics[width=.49\columnwidth]{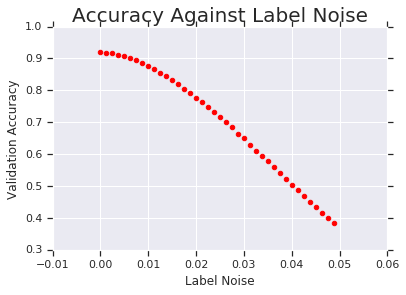}
\includegraphics[width=.49\columnwidth]{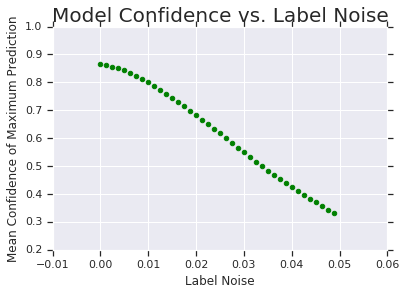}
\caption{\textbf{Top Left:} Calibration measured through three metrics on MNIST with randomly assigned labels, with the fraction of randomly assigned labels on the x axis (all others labels are correct). Models (here, multinomial logistic regression) are re-trained at each noise level. The base level of calibration error will be different, as ACE and SCE account for all predictions, while ECE only accounts for the prediction with maximum confidence for each sample. These predictions have different average calibration errors.\textbf{Top Right:} Often, important predictions (confidence > 0.1) are not captured in the calibration error by ECE. As models trained with more label noise exhibit more uncertainty, that number increases as the amount of label noise increases. \textbf{Bottom:} As label noise increases, model output certainty and accuracy decrease.}

\end{figure}

\begin{figure}
\includegraphics[width=.9\columnwidth]{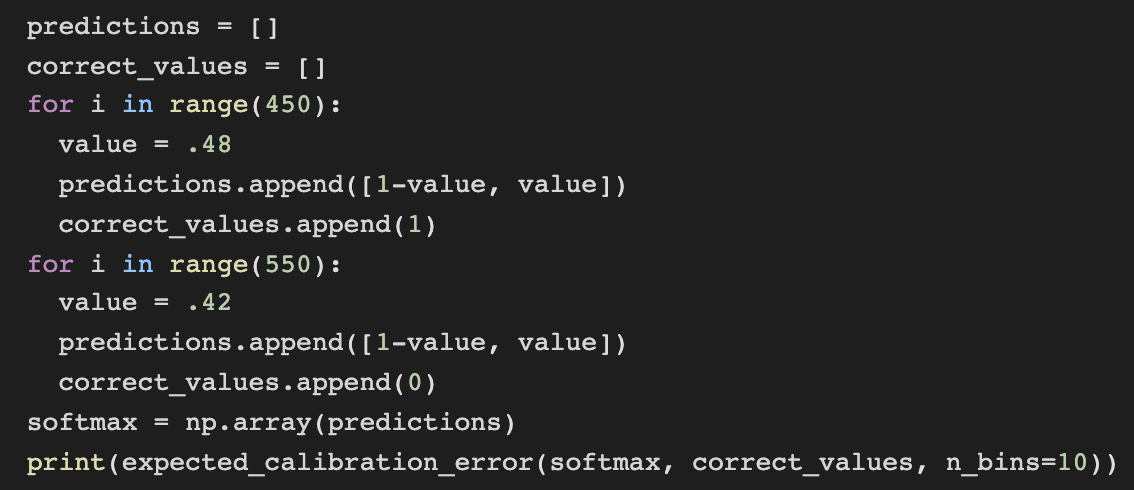}
\caption{This python code (with numpy imported as np, and with an expected calibration error function loaded in the environment) implements a pathology in ECE which will return 0 calibration error by including overconfident and underconfident values in the same bin. The softmax here has 450 predictions at .52 (ECE only looks at the maximum value) and evaluates all of them to be incorrect, despite being predicted with high confidence. The softmax also has 550 predictions at .58, all evaluated to be correct, though they're predicted with low confidence (relative to 1 for correct and 0 for incorrect). Despite this extreme miscalibration, the expected calibration error is exremely low (~0.003).}
\end{figure}

\section{Label Noise}

We train softmax regression (multinomial logistic regression) models at $40$ levels of label noise from $0.0$ to $0.05$. The label noise value corresponds to the percentage of labels that are randomly assigned, where the correct label is included as one of the possible random assignments.

One way to measure the importance of a sample is its confidence. We look at the fraction of important predictions ECE omits by setting a threshold confidence ($.01$ in our visualized example, see Figure in appendix Top Right) and observe that this fraction rises with the difficulty of the modeling task (measured by model accuracy and mean maximum confidence, see Figure 1 Bottom). This result holds for a wide range threshold levels.

\section{Calibration Metric Labeling}

Order is: binning scheme, max probs, class conditional, threshold, norm\\
0 ('even', True, True, 0.0, 'l1') \\
1 ('even', True, True, 0.0, 'l2') \\
2 ('even', True, True, 0.01, 'l1') \\
3 ('even', True, True, 0.01, 'l2') \\
4 ('even', True, False, 0.0, 'l1') \\
5 ('even', True, False, 0.0, 'l2') \\
6 ('even', True, False, 0.01, 'l1') \\
7 ('even', True, False, 0.01, 'l2') \\
8 ('even', False, True, 0.0, 'l1') \\
9 ('even', False, True, 0.0, 'l2') \\
10 ('even', False, True, 0.01, 'l1') \\
11 ('even', False, True, 0.01, 'l2') \\
12 ('even', False, False, 0.0, 'l1') \\
13 ('even', False, False, 0.0, 'l2') \\
14 ('even', False, False, 0.01, 'l1') \\
15 ('even', False, False, 0.01, 'l2') \\
16 ('adaptive', True, True, 0.0, 'l1') \\
17 ('adaptive', True, True, 0.0, 'l2') \\
18 ('adaptive', True, True, 0.01, 'l1') \\
19 ('adaptive', True, True, 0.01, 'l2') \\
20 ('adaptive', True, False, 0.0, 'l1') \\
21 ('adaptive', True, False, 0.0, 'l2') \\
22 ('adaptive', True, False, 0.01, 'l1') \\
23 ('adaptive', True, False, 0.01, 'l2') \\
24 ('adaptive', False, True, 0.0, 'l1') \\
25 ('adaptive', False, True, 0.0, 'l2') \\
26 ('adaptive', False, True, 0.01, 'l1') \\
27 ('adaptive', False, True, 0.01, 'l2') \\
28 ('adaptive', False, False, 0.0, 'l1') \\
29 ('adaptive', False, False, 0.0, 'l2') \\
30 ('adaptive', False, False, 0.01, 'l1') \\
31 ('adaptive', False, False, 0.01, 'l2') \\
\section{Challenges in Calibration}

\sidenote{dt: can we remove Section 5-6.2? we may also no longer need an isolated experiments section if it's involved across multiple sections. i vote for moving these challenges + limitations to discussion if space permits}
Before describing new metrics for calibration, we first outline broad challenges with designing such metrics.

\subsection{Ground Truth \& Comparing Calibration Metrics}
There are many challenges in measuring a network's calibration, beginning with the absence of ground truth.
In principle, one can limit comparisons to controlled, simulated experiments where ground truth is available by drawing infinitely many samples from the true data distribution. However, even with ground truth, any estimation property such as bias remains difficult to compare across estimators, as ``ground truth error'' is multi-valued and estimators may make measurements for different elements of these values. Specifically, calibration is a guarantee for all predicted probabilities $p\in[0,1]$ and class labels $y\in\{0,\ldots,K-1\}$ (see also the calibration function of \citet{vaicenavicius2019evaluating}). An estimator may have lower bias and/or variance in estimating the error for specific ranges of $p$ and $y$ but higher bias and/or variance in other ranges.

The other major challenge with respect to ground truth is that due to the differences in their assumptions, different calibration metrics approximate different overall calibration values. ECE doesn’t look at all predictions, and is estimating a different value than SCE. The ACE vs SCE comparison is possible with respect to the probabilities they look at, but as the static binning scheme imposes a different bias than the adaptive scheme, it’s a challenge to compare static metrics to adaptive metrics.

\subsection{Weighting}
Because calibration error is multi-valued, and a desirable metric is scalar-valued, the question of how to weight probability values (where one can see thresholding as a 0/1 weighting on datapoints below / above the threshold) creates a set of differences in calibration error metrics that choose to emphasize different aspects of calibration performance.
In many contexts what matters most is the rare event---the network's classifications leading up to an accident, the presence of a planet, or the presence of a rare disease. 
Knowing the difference between whether a class's true probability is .01 and .001 (the difference between 1 in 100 and 1 in 1000) is both extremely difficult to discern and may be much more relevant than a difference between .3 and .301, which these calibration metrics would treat as equivalent. In these contexts, weighting the ends of the interval close to 0 and 1 would be ideal.

Additionally, in the context of out-of-distribution detection, we would prefer to be well-calibrated in the middle of the spectrum, whereby a prediction of .5 (high uncertainty) is more likely to happen.

\begin{figure}
\includegraphics[width=.35\columnwidth]{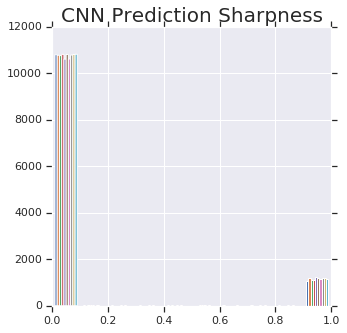}
\includegraphics[width=.49\columnwidth]{figures/ace_100.png}
\caption{\textbf{Left:} Model sharpness, where the vast majority of confidence scores are very near 1 or very near 0, motivates ACE. This class-conditional histogram of confidence scores on 12,000 samples from Fashion-MNIST demonstrates that sharpness. \textbf{Right:} Binning scheme chosen by ACE measured at each step of training on Fashion-MNIST. Each bin measuring the calibration error of datapoints within a particular range has a lower bound to the bin and an upper bound. As the model's predictions become sharper, the chosen bins adapt to focus on regions with a high concentration the predictions. Each color represents the lower bound of the nth largest bin, for 100 bins.}
\end{figure}

\end{document}